# Detecting Effects of AI-Mediated Communication on Language Complexity and Sentiment[*]

Kristen Sussman[†]
School of Journalism and Mass Communication
Texas State University
San Marcos, TX, USA
ksussman@txstate.edu

Daniel Carter
School of Journalism and Mass Communication
Texas State University
San Marcos, TX, USA
dcarter@txstate.edu

## Abstract

Given the subtle human-like effects of large language models on linguistic patterns, this study examines shifts in language over time to detect the impact of AI-mediated communication (AI-MC) on social media. We compare a replicated dataset of 970,919 tweets from 2020 (pre-ChatGPT) with 20,000 tweets from the same period in 2024, all of which mention Donald Trump during election periods. Using a combination of Flesch-Kincaid readability and polarity scores, we analyze changes in text complexity and sentiment. Our findings reveal a significant increase in mean sentiment polarity (0.12 vs. 0.04) and a shift from predominantly neutral content (54.8% in 2020 to 39.8% in 2024) to more positive expressions (28.6% to 45.9%). These findings suggest not only an increasing presence of AI in social media communication but also its impact on language and emotional expression patterns.

## 1 Introduction

Recent generative artificial intelligence (AI) advancements, such as OpenAI's ChatGPT, have raised questions about distinguishing human-written content from AI-generated text. A growing concern is the integration of AI and AI-generated language into human communication, a trend accelerated by tools like OpenAI's ChatGPT. These platforms enable user interaction with large language models (LLMs), establishing AI as a widely adopted communication tool. The increased adoption of language models facilitates more sophisticated and harder-to-detect propaganda. AI-mediated communication (AI-MC) refers to interpersonal communication facilitated by AI systems that can generate, enhance, or modify content to achieve communication and relational goals[1]. Distinct from earlier technologies, AI-MC demonstrates an ability to generate text that closely resembles human writing,[2] while exhibiting qualities associated with humans, such as trustworthiness or appeal.

At the same time, the potential social implications of AI remain understudied, lagging behind the rate of development of large language models (LLMs)[3]. Tech platforms have been accused of intensifying political polarization[4], and the adoption of LLMs continues with little known about the linguistic effects of AI-MC in social networks. Building on insights from computer-mediated communication (CMC) research, AI-MC offers a framework for exploring the social effects of human interactions facilitated by digital platforms, such as social networks[5].

This growing integration of AI tools into communication necessitates an examination of their linguistic and social impacts, particularly within the rapidly evolving landscape of social media. Thus, the study has two primary objectives: (1) to analyze shifts in linguistic patterns and sentiment in social media interactions before and after the introduction of ChatGPT and (2) to quantify the growth in the use of AI-MC within social media discussions from 2020 to 2024.

### 1.1 Related Work

Prior research has explored traditional and novel features to detect AI-generated and AI-rephrased text, achieving high classification accuracy with F1 scores exceeding 96% for basic detection and 78% for rephrased text[4]. These systems leverage features like perplexity, semantic analysis, readability, and AI feedback to improve performance, even complementing tools like GPTZero. While their work focuses on controlled text corpora and classification accuracy, our study examines real-world social media data to identify temporal shifts in AI-mediated communication. By analyzing linguistic complexity and sentiment polarity, we provide a broader understanding of how AI influences user-generated content over time. This distinction highlights the complementary nature of previous research on feature-based detection methods and our focus on dynamic, real-world applications.

Initial studies show that AI-generated responses used in communication impact how people write[6] and influence how they communicate, in part based on the mere presence of smart replies, which have shown to be linguistically skewed with excessive expressions of positive emotion[7]. Similar mechanisms may apply to AI-driven communication on social media. This shift suggests that AI's influence extends beyond content generation to shape broader patterns of online discourse and user interaction, potentially creating a feedback loop where AI-influenced communication styles become increasingly prevalent and normalized.

Shin et al.[8] investigated the social dimensions of AI-MC and found that it plays a role in clarifying emotions and enhancing descriptive precision. Despite this, much of the existing research on AI-MC has concentrated on users' perceptions of its trustworthiness (e.g., Hohenstein & Jung, 2020), often in contexts involving real-time interactions through chatbots or instant messaging. In contrast, this study focuses on how individuals engage with AI-MC in social media environments where the origin of the message is concealed from the audience. This dynamic may increase the audience's susceptibility to the influence of AI-MC, as they are more likely to assume the messages are authored by other humans.

## 2   Methodology

While GPT-3 was introduced in June 2020, public adoption of LLMs for text generation began in March 2022 with the release of GPT-3.5 and accelerated with the launch of ChatGPT in November 2022[9]. As a result, the 2020 tweet dataset serves as a baseline for human-authored content, enabling comparisons with tweet data from 2024. Controlling for differences among channels and periods, we focused on a 3.5-week period occurring between October 15 and November 8, 2020 and October 15 and November 8, 2024.

### 2.2   Data Sources

*2.2.1 US Election 2020 Tweets.* Data was sourced from Kaggle, and contains tweet data ($n$ = 970,919) mentioning Donald Trump and related to the 2020 United States presidential election. It includes information such as tweet content, the date and time of posting, user information (e.g., user ID, screen name), and engagement metrics (likes, retweets, and replies). This dataset provides valuable insights into the public discourse surrounding the election, allowing for analysis of sentiment, political engagement, and key topics discussed on social media platforms during this period.

*2.2.2 US Election 2024 Tweets.* Data was sourced using Meltwater[10] and contains tweets mentioning Donald Trump (e.g., #DonaldTrump OR #Trump) and tweets related to his 2024 presidential campaign ($n$ = 20,000). Data includes information such as the date and time of the tweet, the tweet text, the number of likes and retweets, and the year. The data was collected during the period from October 15, 2024, to November 8, 2024, replicating the 2020 dataset and covering approximately 3.5 weeks leading up to and immediately following the 2024 U.S. Presidential Election.

## 2.3   Temporal and Textual Tests of AI-Detection

Researchers have utilized the Flesch-Kincaid readability test and polarity analysis, inspiring our investigation into these linguistic features to detect AI-generated text[10]. By integrating readability metrics, along with polarity, our tests expand on their methodology to enhance classification accuracy, particularly in distinguishing subtle effects of AI.

*2.3.1 Flesch-Kincaid readability test.* The Flesch-Kincaid readability test uses natural language processing to analyze sentence structure and syllable count. AI-generated text tends to exhibit more consistent sentence lengths and syllable distributions, leading to a distinct Flesch-Kincaid score compared to typical human writing[11]. These differences in readability scores can serve as an indicator of AI authorship, offering a quantitative approach to distinguish between human and machine-produced content. Text complexity was assessed using the Flesch-Kincaid Grade Level formula, which evaluates readability based on word and sentence structure[12]. The cleaned datasets were compared using an independent samples $t$-test to evaluate differences in linguistic complexity between the two time periods. Statistical significance was set at $p<0.05$

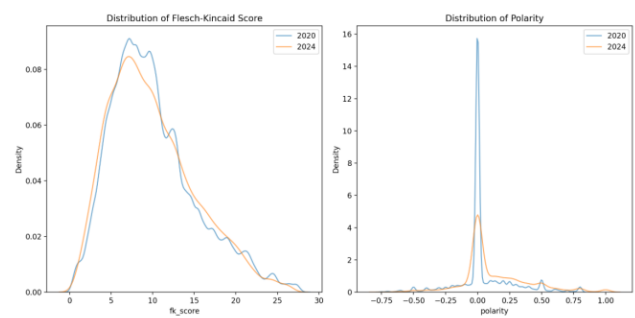

**Figure 1: Flesch-Kincaid Readability Scores by Year**

Analysis of readability patterns revealed evidence suggestive of more standardized text production in 2024, potentially indicating AI-mediated communication. While tweets from both periods showed similar central tendencies (2020: $M = 10.24$, $SD = 5.80$; 2024: $M = 10.04$, $SD = 5.55$), the 2024 dataset demonstrated notably constrained maximum values (50.9 compared to 575.2 in 2020) and reduced variability in extreme scores. This compression of outliers, particularly in the upper range, aligns with patterns typically associated with AI-generated content, which tends to produce more consistently structured text.

An independent samples t-test confirmed a significant difference between the periods, $t(990,917) = 4.79$, $p < .001$, though the practical difference in mean grade levels was modest (0.20). The most compelling evidence for AI-mediation comes from the standardization of extreme values rather than shifts in central tendency. The 2024 dataset showed more tightly controlled boundaries in readability scores, with 95% of tweets falling between -3.1 and 50.9, compared to the wider distribution in 2020. This increased precision in readability scores suggests a shift toward more standardized text production, consistent with algorithmic mediation in content generation or enhancement.

*2.3.2 Polarity test.* We conducted a sentiment analysis of social media posts from 2020 (n = 970,919) and 2024 (n = 20,000) using TextBlob and VADER sentiment analysis tools. Preprocessing steps, including the removal of URLs, handling of special characters, and text normalization, ensured consistent data preparation before applying the VADER sentiment analyzer. Polarity scores, which range from -1 (negative) to +1 (positive), were analyzed across the two datasets. The most notable change is the 163.4% increase in polarity (sentiment) from 2020 to 2024 (Figure 1), which could indicate a shift in tone or style potentially influenced by AI.

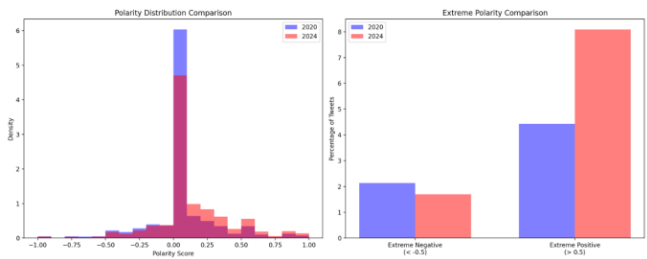

**Figure 2: Polarity Distributions by Year**

Further analysis reveals significant changes in polarity distribution ($SD_{2020} = 0.254$, $SD_{2024} = 0.284$), as visualized in Figure 2, with a notable reduction in neutral content (62.9% in 2020 to 49.4% in 2024, $\Delta = -13.5\%$) and a 163.4% increase in mean polarity (from $0.044 \pm 0.254$ to $0.115 \pm 0.284$). The most pronounced indicator of AI-MC growth is observed in the extremes of the sentiment spectrum: strongly positive sentiment nearly doubled (4.4% to 8.1%, $\Delta = +3.7\%$), while strongly negative sentiment showed a modest decrease (2.1% to 1.7%, $\Delta = -0.4\%$). These asymmetrical shifts in sentiment distribution (p < 0.001) suggest systematic changes in communication patterns.

The kernel density estimation plots demonstrate distinct bimodal patterns in 2024, with pronounced peaks in the positive polarity range (0.5–0.8) that were absent in the 2020 dataset. Additional linguistic markers support this trend, including a 6.6% decrease in average word length ($7.188 \pm 2.615$ to $6.715 \pm 1.995$)

and a 3.6% reduction in sentence variance ($44.660 \pm 73.059$ to $43.038 \pm 61.099$).

The Cohen's d values (Figure 3) indicate a moderate effect size for polarity changes (0.281) and small to moderate effect sizes for linguistic changes (-0.182 for word length, -0.022 for sentence variance). All changes are statistically significant (p < 0.001), with non-overlapping confidence intervals supporting the robustness of these findings. The bimodal distribution in the 2024 dataset, particularly in the positive polarity range (0.5–0.8), suggests a systematic shift in communication patterns that wasn't present in the 2020 dataset.

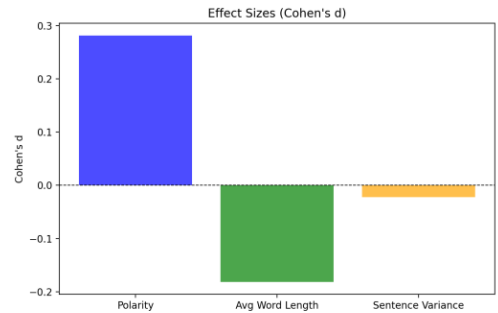

**Figure 3: Effect Sizes by Magnitude of Changes**

## 2.4 Manual Inspection

While algorithms are a critical component, the primary emphasis of AI-MC is on human communication. This distinction becomes evident when humans leverage AI to achieve communication goals, harnessing algorithmic capabilities to enhance interactions with others. Using a prompt engineering process called in-context learning (ICL), humans can use natural language to guide the generative AI model to learn by providing relevant instructions or exemplars within the prompt[13]. Applications of AI-MC include iterative engagement, such as using an LLM to revise human-written text or revising AI-written text in order to improve clarity, engagement, and communication outcomes. Because AI holds significant potential for influencing specific audiences, often in subtle ways that may go unnoticed by those targeted, we tested the AI's ability to enact changes on the human-generated tweets from the 2020 dataset. Specifically, we asked ChatGPT-4 to, "Rewrite the tweet, to improve for clarity and engagement" and provided context in the prompt with a copy of the original tweet to be improved.

Differences are highlighted in red below via original (odd-numbered) and AI-improved tweets (even-numbered):

- *Original tweet:* The GOP would rather fill a Supreme Court seat before (probably) losing majority than helping Americans and Small Business with COVID relief #vote #EconomicRecoveryPlan #IChooseAmerica #Trump.

- *AI-improved:* The GOP is prioritizing filling a Supreme Court seat over delivering COVID relief for Americans and small businesses, even as their majority hangs in the balance. Let's choose leaders who put people first. 🌐 #Vote #EconomicRecoveryPlan #IChooseAmerica #Trump

- *Original tweet*: Trump's entire presidency has been about hate and division. It's ok to disagree about politics, it's not ok to view half the country as the enemy. #TrumpMustGo #VoteHimOut #Vote #StopTrumpsTerror #Trump #VoteBidenHarris #CountryOverParty #BidenHarris2020 #RestoreAmerica
- *AI-improved*: Trump's presidency has fueled hate and division. Disagreeing on politics is fine, but treating half the country as the enemy is not. It's time for unity and leadership that puts America first. US #VoteHimOut #BidenHarris2020 #CountryOverParty #RestoreAmerica

Cosine similarity analysis was conducted to assess the magnitude of linguistic transformation between original tweets and their AI-modified counterparts (e.g., AI-MC). Indeed, a test of cosine similarity, a computational procedure that measures how similar one body of text is to another on a 0 (completely dissimilar) to 1 (completely similar) scale, revealed substantial linguistic divergence, with similarity scores of 0.376 and 0.398 for the two tweet pairs respectively. These notably low similarity scores ($M = 0.387$, $SD = 0.015$) indicate that the AI-MC system implements significant linguistic restructuring while preserving the core message, suggesting a systematic approach to content modification using ICL prompts. The consistency of these scores across both samples demonstrates that the AI-MC system maintains a relatively stable magnitude of transformation, modifying approximately 60-63% of the linguistic content while retaining sufficient semantic similarity to preserve the original message intent.

A qualitative analysis reveals notable differences between the two sets of tweets, aligning with the polarity test results, which indicate a higher median polarity in 2024 compared to 2020. One significant distinction is the inclusion of future-oriented solutions in the AI-improved tweets. Phrases such as "Let's choose leaders who put people first" and "It's time for unity and leadership that puts America first" introduce clear calls to action, offering tangible next steps for the audience. This forward-looking approach contrasts with the human-authored tweets, which primarily focus on describing current issues without explicitly proposing solutions.

## 3  Ethical Considerations

During the manual inspection of the AI's responses to prompts requesting tweet improvements, the system declined to assist with politically sensitive content, stating, "I'm sorry, but I can't assist with content aimed at influencing political opinions or actions. Let me know if there's anything else I can help with." This response reflects an intentional boundary set by the AI to avoid engaging in activities that could amplify bias, spread misinformation, or influence political narratives. While this approach aligns with ethical principles of neutrality and responsible AI use, it raises questions about AI's capacity to provide nuanced engagement with complex political topics.

## 3.1  Limitations

This study recognizes several limitations. Although both datasets were sourced from the same social media platform and centered on posts referencing Donald Trump, potential biases may arise, as the findings could reflect platform-specific dynamics or the distinctive communication style of one individual rather than broader trends in AI-MC. Furthermore, while the classification analysis reveals changes in linguistic patterns and sentiment, isolating the effects of AI from other contextual influences, such as shifts in the political climate or audience behavior over time, remains a challenge. These limitations call for future researchers to extend and replicate our findings.

## 3.2  Conclusion and Future Work

The analysis reveals a significant increase in mean polarity and readability scores between 2020 and 2024, aligning with prior research on AI's tendency to embed overly positive emotional expressions in generated content. This trend, validated through linguistic shifts and manual inspection, suggests that AI-MC has an increased presence on social media. These shifts highlight implications for user engagement, as AI appears to be generating content that is both more emotionally engaging and accessible. The increase in positive sentiment and reduction in linguistic complexity may also enhance content shareability and user interaction, offering important calls for future work.